\documentclass[11pt,dvips,twoside,letterpaper]{article} 
\usepackage{times,fancyhdr}
\usepackage{graphicx}
\usepackage{geometry}
\usepackage{titlesec}
\usepackage{hyperref}
\titlelabel{\thetitle.\quad}                                

\usepackage[labelsep=period]{caption}                       


\makeatletter 
\def\cleardoublepage{\clearpage\if@twoside \ifodd\c@page\else%
    \hbox{}%
    \thispagestyle{empty}
    \newpage%
    \if@twocolumn\hbox{}\newpage\fi\fi\fi} 
\makeatother

\usepackage{amsmath}
\usepackage{amsfonts}
\usepackage{amsthm}
\usepackage{amssymb}

\begin{document}

\title{Computational Theories of Curiosity-Driven Learning}


\author{Pierre-Yves Oudeyer\thanks{\url{http://www.pyoudeyer.com}}\\
Inria and Ensta ParisTech, France}
\date{}
\maketitle
\thispagestyle{empty}
\setcounter{page}{1}




\begin{abstract}
What are the functions of curiosity? What are the mechanisms of curiosity-driven learning?
We approach these questions about the living using concepts and tools from machine learning and developmental robotics.  We argue that curiosity-driven learning enables organisms to make discoveries to solve complex problems with rare or deceptive rewards. By fostering exploration and discovery of a diversity of behavioural skills, and ignoring these rewards, curiosity can be efficient to bootstrap learning when there is no information, or deceptive information, about local improvement towards these problems. We also explain the key role of curiosity for efficient learning of world models.  We review both normative and heuristic computational frameworks used to understand the mechanisms of curiosity in humans, conceptualizing the child as a sense-making organism. These frameworks enable us to discuss the bi-directional causal links between curiosity and learning, and to provide new hypotheses about the fundamental role of curiosity in self-organizing developmental structures through curriculum learning. We present various developmental robotics experiments that study these mechanisms in action, both supporting these hypotheses to understand better curiosity in humans and opening new research avenues in machine learning and artificial intelligence. Finally, we discuss challenges for the design of experimental paradigms for studying curiosity in psychology and cognitive neuroscience. 
\index{curiosity} \index{intrinsic motivation} \index{world models} \index{free-energy principle} \index{learning progress hypothesis} \index{lifelong learning} \index{predictions} \index{predictive processing} \index{artificial intelligence} \index{machine learning} \index{robotics} \index{robotics!developmental robotics} \index{development} \index{developmental psychology} \index{curriculum learning} \index{self-organization} \index{computational modeling} \index{models of curiosity} \index{theory} \index{theory!computational theories} \index{curiosity-driven learning}
\end{abstract}



\section{Introduction}
Humans and many other animals spontaneously explore their environments. This often happens without a pressure for finding extrinsic rewards like food, and without external incentives from their social peers.
Such spontaneous exploration seems to be produced by an internal mechanism pushing them to make sense of their world: they explore for the intrinsic purpose of getting better at predicting and controlling the world. This spontaneous investigation of the environment, and of the link between one's own physical and cognitive capabilities and the environment, can take many different forms. This ranges from babies trying various ways to locomote, or exploring grasping, manipulating, banging or throwing of all sorts of objects, or testing how social peers respond to vocalizations, to children practicing tool building with wooden sticks, or throwing wooden sticks in rivers to see how they will flow, to adults searching information about a hobby, learning a new sport, or a scientist turning his telescope towards faraway galaxies.
	
\pagestyle{fancy}
\fancyhead{}
\fancyhead[EC]{\it Pierre-Yves Oudeyer}
\fancyhead[EL,OR]{\thepage}
\fancyhead[OC]{\it Computational Theories of Curiosity-Driven Learning}
\fancyfoot{}
\renewcommand\headrulewidth{0pt} 

	All these exploratory behaviours can be seen as questions posed by the organism about its environment or about the relation between its environment and it's own current state of knowledge and skills. These questions can be formulated in various ways ranging from actual physical/behavioural experimentation to formulating a linguistic question.
 
\index{play} \index{play!free play} \index{complexity}
These mechanisms have been discussed from various perspectives in the scientific literature, and in particular using the related concepts of curiosity \cite{berlyne1960a}, intrinsic motivation \cite{harlow1950a}, and free play \cite{bruner1976a}. Across many different fields, theorists have suggested that “interest” is engaged by what is just beyond current knowledge, neither too well known nor too far beyond what is understandable.  This idea has been offered many times in psychology, through concepts like cognitive dissonance \cite{kagan1972a}, optimal incongruity \cite{hunt1965a}, intermediate novelty \cite{berlyne1960a,kidd2012a} and optimal challenge \cite{csikszenthmihalyi1991a}, and recent research in neuroscience is now investigating the associated neural mechanisms \cite{gottlieb2013a}.

Several formal frameworks have recently enabled to improve our theoretical understanding of these mechanisms. This includes frameworks considering the curiosity system as a machine which goal is to build predictive and control models of the world \cite{kaplan2007a,oudeyer2016a} - and sometimes the brain as a whole is conceptualized like this as in the free energy principle \cite{friston2017a,friston2017b}. Related to this, reinforcement learning and optimization frameworks consider curiosity as a mechanism which allows to generate diversity in exploration, enabling to get out of local minima in searching for behaviours that maximize extrinsic rewards or fitness \cite{schmidhuber1991a,barto2013a,baldassarre2013a,lehman2008a,bellemare2016a,forestier2017a,colas2018gep}. 

\section{Curiosity for Exploration and Discovery in an Open World}
\index{discovery} \index{novelty} \index{learning} \index{reinforcement learning} \index{rewards} \index{rewards!rare rewards} \index{intrinsic rewards} \index{evolution} \index{evolution!evolutionary benefits of curiosity} \index{exploration!spontaneous}
 
An apparent evolutionary mystery is that such spontaneous investigations of the environment are often very costly in time and energy, and do not appear at first sight to provide a direct benefit for feeding, mating, survival and reproduction. So how could such mechanisms evolve? What is their function? 

In an uncertain world with rare resources, one could expect that organisms spare their energy to explore only parts of the environment that are likely to provide information about where to get these resources. However, the real world is not only uncertain, but from the point of view of many basic physiological needs like finding food, it is full of multimodal multidimensional stimuli that are not obviously relevant to these needs. Animals have initially little ideas of what kinds of actions are required to fulfil them. Thus, when extrinsic rewards (resources) are hard to find, the main challenge is not to estimate the relative reward probabilities associated to a few reward-relevant options in order to maximize the efficiency of a known solution. Rather, the challenge is to discover the first few bits of rewards and how to build coarse strategies to get them. In this context where discovery of new strategies and new outcomes becomes the main issue (as opposed to refining a known strategy for getting a known outcome), one can better see how to make sense of curiosity-driven exploration in living organisms.

Let's take the example of an 8-9 months old baby, sitting on the ground and alone in a room. He has seen a box of sweets on top of a kitchen's furniture, and aims to get them. While the baby might really be motivated to get the sweets for a moment, this task is for him a real conundrum. The situation is full of multiple kinds of deep uncertainties, and most importantly deep unknowns. First of all, the child has only a very approximate idea of the current state of the world: he sees the sweet box from far away with his eyes, and estimating simple things such as distance and height is already very difficult, since he is mostly used to interact with objects that are already in his hands, and has limited skills to estimate the state of far away objects. Second, the child has no clue about how to get to the sweet box, and has no clue where to look to find information about a solution. He does not even know how to stand up on its two feet, and his crawling strategy is very imprecise to move around. He does not know yet what a tool is, and his brain cannot even imagine at this point the possibility to push a chair next to the furniture, then try climb it to get to the box (at this point, chairs are perceived as obstacles for him). Here, he is much beyond uncertainty: it is meaningless for him to compute probabilities or uncertainties associated to the success of this strategy (and its associated sub-goals), as they involve events that are not already part of the space of hypotheses he can consider. Just think of the intermediate skill of standing up on its two feet and climbing the chair. Even if targeting these skills can be suggested by observing its social peers, they involve such complex internal muscular synergies that initially the child has also little cue about what patterns of muscle activations, and what proprioceptive and external visual information to attend to control these skills. Also very little can be inferred from observing others, as high-dimensional muscular activations and covert attention in these skills are not directly observable. 

So what could the child do? He might consider a proxy internal measure for guiding its search for the sweet box, such as the current distance between his body and the sweets. Yet, optimizing it with its current skills would just make it crawl to the bottom of the furniture and extend its arm for a hopeless far reach. In that case this proxy measure would be less sparse than the sweet reward itself, but it would be highly deceptive, and equally inoperant. Rather, a good strategy for the child shall be to simply forget about this target sweet box for a while, go back to his playground and explore spontaneously, driven by curiosity, various forms of skills ranging from body locomotion to object manipulation and tool use. Playing around with its own body shall then lead him to discover both the concept of and strategies for climbing up furniture, as well as the concept and strategies for using objects of all kinds as tools. Then, when coming back a few months later to the task of getting to the sweet, and once he will have acquired a skill repertoire including walking, pushing chairs around, and using them as tools to get upwards, the solution might become much more accessible. At this point only the problem will be looking like a classical reinforcement learning problem in the lab with few obvious relevant options to choose from. 

Research in developmental robotics and artificial intelligence has confirmed this intuition in the last decade through computational and robotics experiments. Several strands of works have considered the problem of how a machine could solve difficult problems with rare or deceptive rewards \cite{barto2004a,schmidhuber1991a,lehman2008a}, sometimes directly aiming at modelling how human children explore and learn in these contexts \cite{oudeyer2007a,oudeyer2016a}. They found that various forms of curiosity-driven exploration can indeed be the key to make discoveries and solve these complex reinforcement learning problems. 

\begin{figure}[!ht]
\begin{center}
\includegraphics[width=0.6\columnwidth]{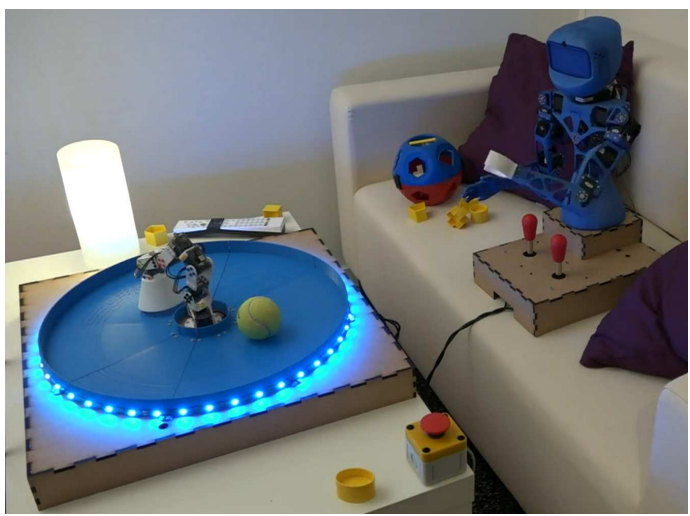}
\end{center}
\caption{\textbf{Curiosity-driven exploration through autonomous goal setting and self-organized curriculum learning} in the experimental setup presented in \cite{forestier2017a} (see video: \url{https://www.youtube.com/watch?v=NOLAwD4ZTW0}). Here a humanoid robot is surrounded by objects. The robot can learn to control some of them, while others are unlearnable distractors.  Some objects can interact with each other, e.g. be used as tool to control other objects. The robot initially does not know which skills and objects are learnable, nor that objects may interact with each others. If an engineer would like the robot to learn how to move the ball forward by providing rewards only when the ball moves, then using classical reinforcement learning approaches would fail. Indeed, RL approaches combine hill-climing mechanisms that require observations of non-zero reward to improve the current solution, and rely on random exploration otherwise. Here, the time required by random exploration to enable the robot to find such a rare reward would be prohibitively long. The robot initially has no idea where to look for cues to solve this task. Indeed, moving the ball entails moving the white electric crane toy, which can only be moved by pushing one of the two joysticks, which can itself only be moved by appropriate movements of the hand, requiring specific sequences of motor commands in the joints. A more efficient approach is to use curiosity-driven exploration, where one lets the robot self-generate its own goals for controlling various objects, and spend time on the ones which produce maximal learning progress. The internal generation of goals, and the focus on goals providing maximal learning progress, model a form of curiosity. Doing so, the robot will first focus on playing with its hand, which is initially providing maximal learning progress. Then, it will discover as a side effect how to move the joysticks. In turn, due to the physical couplings in the environment, focusing on learning about the joysticks makes the robot discover how to move the crane toy and the ball in a few hours.}
\label{fig:1}
\end{figure}

An example is the artificial intelligence and robotics experiment presented in \cite{forestier2017a} (see Figure~\ref{fig:1}), where a high-dimensional humanoid 'baby' robot is situated in an environment with many objects. Some of them are more or less difficult to learn to control, and some other are uncontrollable by the robot, such as another robot moving around. However, the 'baby' robot does not know initially which objects it can learn to control and which ones it cannot. Among these objects, a ball is placed in an area beyond the direct reach of the robot. Yet, other objects can be used as tools to enable the robot to move the ball. The robot can use its hand to push a joystick, which in turn moves a tele-operated crane toy, and this crane toy can push the ball. However, the robot has no concept of tool initially and does not know that these objects can physically interact: it has no pre-coded way to represent such physical interactions. It can send low-level motor commands into the motors of its arm. It can perceive object positions and movements individually. Yet, it does not know initially how specific sequences of motor commands relate to potential movements of each object (and how the movements of objects might relate to each other). While this robotic situation is simplified as compared to most real world situations encountered by human infants, it is already extremely complex. Indeed, the sequence of motor commands for a simple arm movement needed to reach for an object amounts to specifying several dozen continuous numbers, while the perception of the movement of each single object during one second is also encoded by several dozens continuous numbers. As a whole, the sensorimotor space that the robot explores has several hundred dimensions. 

Given such an environment, let us imagine an engineer imposing the following external objective to the robot: it should learn to move the ball forward at a maximal speed. To define this objective, the engineer can design an extrinsic reward signal: each time the robot tries a sequence of motor commands that produce a movement of the ball, this reward is a scalar number proportional to the speed and target direction of the ball. Each time the motor commands do not produce any ball movement, the reward is zero. 

The standard machine learning approach to enable the robot to solve this problem is reinforcement learning (RL) \cite{sutton1998reinforcement}. This can be viewed as a family of optimization algorithms that learn optimal controllers, i.e. learn to produce sequences of motor commands that maximize the reward. The way standard approaches to RL work is through a combination of hill-climbing (gradient descent) and random exploration. For example, state-of-the-art deep reinforcement learning algorithms for learning continuous control, such as DDPG and related algorithms \cite{lillicrap2015continuous,schulman2017proximal,sigaud2018policy}, work by alternating between updating the current controller solution in order to climb the hill of rewards (this requires that rewards of different magnitudes are observed when slightly changing the controller), and producing random perturbations of the current best controller to obtain further information about the reward distribution. 

However, such an RL approach will not work in the robotic environment considered here. Indeed, in this particular environment, the problem of moving the ball forward is said to be a 'rare reward' problem. Indeed, due to the structure of the environment, the vast majority of possible sequences of motor commands will produce a reward of zero. Actually, most random sequences of motor commands will make that the robot will not even touch the joystick in front of him. Thus, if the robot tries actions randomly, there is a very low probability to get a non-zero reward, i.e. that its arms touches the joystick in a peculiar way that makes the crane toy in a peculiar way that makes the ball move. This is a problem as the hill-climbing mechanisms of RL algorithms need to observe distributions of non-zero reward to update the controller, and doing random exploration of motor commands would here take a prohibitively long time before the first non-zero rewards are found, i.e. before finding the first few action sequences that produce ball movement. This problem of RL approaches that focus on hill-climing of the extrinsic reward is now well-known, and applies to man environments with rare or deceptive rewards\footnote{Deceptive rewards are local improvement of rewards that push the learner to update the controller in wrong directions in relation with the global optimum. For example, in the infant example above, this would be the child observing that elevating the hand towards the sweet box on the table decreases the distance between him and the sweetbox. If the child is considering the hand-sweet distance as a measure of reward, this would be reinforcing this strategy. However, this would get the child away from more efficient strategies on the long term such as fetching a chair and climbing it.} \cite{bellemare2016a,sigaud2018policy,colas2018gep}. 

An alternative computational approach studied in \cite{forestier2017a} has consisted in equipping the robot with the capability to self-generate and explore many other goals than moving the ball. For example, the system can self-generate and focus on goals such as moving the hand in various directions, pushing any of the joysticks along particular trajectories, or trying to move the crane toy in diverse ways. The idea is that the system decides to spend time learning about one of these goals based on an intrinsic measure of the learning progress towards solving them, and ignoring completely how much they provide information about the goal of the engineer (move the ball). Formally, self-generating a goal (e.g. move toy in direction $d$) amounts to self-defining an internal reward function which quantifies how well sequence of motor commands produce the targeted outcome (e.g. the reward can be a scalar value proportional to the difference between the reached toy direction and the target toy direction). Then, the system can use RL algorithms to learn controllers for these internally defined goals and rewards. This whole process models several aspects of curiosity-driven exploration. First, it includes internal spontaneous generation of goals. Second, it includes a meta-cognitive mechanism to assess the relative 'interestingness' of self-generated goals, based on quantifying expected learning progress. This is used by the robot to decide on which goal to focus on at any given moment in time, and to self-organize a curriculum of learning. 

Intuitively this may appear to be an even more difficult situation for the robot, as it consists in providing it with many other potentially difficult problems in addition to the initial problem of the engineer. However, this turned out to facilitate the acquisition of skills to move the ball forward. Indeed, due to the structure and richness of the physical environment, the robot manages to find continuously goals where learning can happen efficiently due to dense reward feedback, leading to an increase in behavioural diversity, and then to an increase of probability to find sequences of motor commands that solve other goals with rarer rewards. In this particular case, this curiosity-driven exploration mechanism pushed the robot to first focus on moving its hand around, leading to a greater diversity of motor behaviours. As a side effect, this enabled the discovery of how to move the joystick \footnote{When the robot targets a goal, but observes effects relevant to another goal as a side effect, it is able to learn retrospectively about this other goal using the collected observation. For example, if it targets to move the hand on the right, but in practice moves it on the left and moves the joystick, it uses this observation to learn how to move the ball left and how to move the joystick. This is a central property of these intrinsically motivated goal exploration processes \cite{baranes2013a, forestier2017a}, which is shared with other related multi-goal learning algorithms in RL such as Hindsight Experience Replay \cite{andrychowicz2017a}.}. Then in turn, moving the joystick became interesting as a source of learning progress. This increased the focus on practicing goals related to the joysticks, which made the robot discover that the crane toy can be moved around, leading quickly to the discovery of how to move the ball. As the first ball movements are discovered, choosing the goal of moving the ball becomes easier, as the hill-climbing mechanism of reinforcement learning can efficiently compute how to change the controller to improve ball movements. Following this process, curiosity-driven exploration enables the robot to learn in a few hours how to solve the task of moving the ball around, even without initially pointing this task as particularly relevant among many other tasks that are also learnt. On the contrary, this would have been prohibitively long and difficult to learn by only considering and focusing on this task externally defined by the engineer. 
\index{curriculum learning}

\index{algorithms} \index{machine learning} \index{deep reinforcement learning} \index{reinforcement learning!intrinsically motivated reinforcement learning}
This example relies on computational models of curiosity-driven exploration that were explicitly motivated by modeling mechanisms of human spontaneous exploration and their role in explaining the discovery of tool use \cite{forestier2016a,forestier2016b}, vocalization \cite{moulin-frier2014a} and language \cite{forestier2017a} by children. However, several other strands of research in artificial intelligence have converged to the same conclusion that curiosity-driven exploration could be a key to solve complex tasks with rare or deceptive rewards. For example, several works in the field of evolutionary computation have shown that novelty search could be essential in solving difficult optimization problems \cite{lehman2008a}, sometimes in combination with the task-specific reward \cite{mouret2012a}, but also sometimes by completely forgetting this task specific reward in the novelty search process \cite{lehman2008a}. In the domain of reinforcement learning, the idea of exploration bonuses \cite{andreae1978a,sutton1990a,dayan1996a} has also been shown to increase the efficiency of reinforcement learning algorithms, with the addition of a reward component measuring quantities such as the novelty of a state \cite{sutton1990a}, prediction progress \cite{schmidhuber1991a,kaplan2003a}, density of state exploration \cite{ostrovski2017a}, predictive information \cite{martius2013a}, predictive information gain \cite{little2013a}, or empowerment \cite{salge2014a}.  Within an intrinsically motivated multi-goal reinforcement learning perspective, measures of competence progress towards self-generated goals were used to automatically generate learning curriculum for acquiring complex skills \cite{baranes2013a,oudeyer2013a,nguyen2014a,forestier2017a}.  Within a hierarchical reinforcement learning perspective, intrinsically motivated RL approaches have also been used as a framework enabling the discovery of hierarchically reusable skills for boosting exploration \cite{barto2004a,kulkarni2016a,machado2017a}. Recent breakthrough in deep reinforcement learning have leveraged these various strands of works. For example, several deep reinforcement learning algorithms were shown to be able to solve difficult problems with rare rewards (e.g. video games), sometimes by even ignoring the score measure \cite{pathak2017a}, either by introducing an intrinsic reward measuring forms of novelty or learning progress \cite{ostrovski2017a,kulkarni2016a}, or by introducing auxiliary tasks \cite{jaderberg2016a,andrychowicz2017a,florensa2017a,cabi2017a} in ways that are related to intrinsically motivated goal exploration approaches in developmental robotics \cite{baranes2013a,forestier2017a,pere2018a}. 

\section{The Child as a Sense-Making Organism}
\index{world models} \index{sense-making} \index{free-energy principle} \index{goals} \index{goals!self-generated goals}

This computational perspective provides a conceptual framework to understand how various forms of curiosity-driven exploration can be instrumental for an organism to discover solutions to extrinsic tasks that are essential to its survival, such as finding extrinsic rewards like food. In this context, it becomes possible to make sense of curiosity-driven exploration in an evolutionary perspective \cite{smith2013a}, and indeed further computational models have shown that intrinsic reward systems could be the result of evolutionary processes optimizing for an individual fitness to reproduce in changing environments \cite{singh2010a}. This perspective converges with the hypothesis proposed in psychology that humans might be equipped with dedicated motivational neural circuitry pushing them to explore the world for the pure sake of making sense of it, and independently (yet complementarily) from the search of extrinsic rewards \cite{berlyne1960a}. For example, Gopnick \cite{gopnik2012a} has suggested that the child could be viewed as a curiosity-driven little scientist equipped with an intrinsic urge to discover the underlying causal structure of the world (see also \cite{chater2016a}). 

Several theoretical models have considered mechanisms of curiosity-driven exploration as organized exploration of the world with the purpose to build good predictive and control models of the world dynamics \cite{oudeyer2007a,friston2017a}. For example, Friston and colleagues \cite{friston2017a} have developed a normative theoretical framework, termed the free energy principle, to characterize the theoretical properties of an ideal optimal Bayesian learner trying to build a good predictive model of the world. More precisely, the free energy principle framework views the brain as an active Bayesian inference system that aims to minimize future expected surprise. A useful property of this framework is to mathematically formalize various forms of uncertainties which are necessarily encountered by the active inference system, and which reduction can lead to corresponding various forms of exploration akin to curiosity (which are also present in the learning progress framework detailed below). For example, sources of uncertainty that appear when mathematically decomposing the expected free energy include uncertainty about hidden states given an action policy: minimizing it leads to active sensing and perceptual inference, i.e. a form of curiosity aiming to improve the subjective estimation of the current world state. Another dimension is uncertainty about policies in terms of expected future states or model parameters: this leads to forms of epistemic exploration and learning, i.e. a form of curiosity aiming to improve the predictive model of what will be observed in the future depending on one's own actions. Yet another dimension is uncertainty about the model structure itself: this leads to structure learning and insight, a form of curiosity aiming to find new abstractions with structures that enable better predictions of what is happening in the environment. While the concept of goals is not directly covered by the free-energy principle, other frameworks like the autonomous goal setting paradigm \cite{oudeyer2007b,forestier2017a}, presented in the robotic experiments above, point to other forms of uncertainty and curiosity centered around goals. Indeed, another form of uncertainty is related to the self-evaluation of goal competences: minimizing these forms of uncertainty lead to curiosity-driven self-generation and practice of goals to learn about one's own competence to achieve them. 

\section{Normative or Heuristic Accounts? The Learning Progress Hypothesis}
\index{theory!normative} \index{theory!heuristics} \index{learning progress hypothesis} \index{theory!free energy principle} \index{theory!learning progress hypothesis}

As the landscape of computational/mathematical theories of curiosity-driven exploration quickly develops \cite{baldassarre2013a,oudeyer2016a}, one becomes better equipped with formal and experimental tools to conceptualize better what curiosity might be, and what it could be used for in humans. However, this landscape of theories also raises multiple open questions to explain human curiosity. A first fundamental question is how to articulate normative approaches (e.g. the free-energy principle, empowerment) with heuristics-based approaches (e.g. the learning progress hypothesis \cite{kaplan2007a,oudeyer2016a}) to account for human behaviour and brain mechanisms. Taking the perspective of David Marr's levels of analysis for understanding cognition and the brain, shall normative approaches be standing at the computational level (describing which are the problems and the quantities that the brain actually target) and heuristic approaches at the algorithmic level (describing which algorithms can actually solve these problems) on top of the implementation level investigating how neural circuitry could implement these algorithms? The answer to this epistemological question does not appear to be resolved yet.

Indeed, while normative approaches like the free-energy principle have been used successfully to account for several behavioural and neural observations ranging from saccadic eye movements to habit learning to place cell activity \cite{friston2017a,friston2017b}, it relies on assumptions that are still speculative about what could be happening in the brain. First, it relies on the assumption that the human brain is capable to achieve (approximate) hierarchical Bayesian inference, but several strands of work have shown a number of situations where the brain uses heuristics that are far away from a Bayesian behaviour (e.g; \cite{gigerenzer1996a}). Second, such a normative Bayesian framework requires that modelled human subjects initially know the full space of possible hypotheses about possible world states, possible policies, possible model parameters, and possible model structures. This deviates strongly from the deep unknown encountered by children as described in the example above: new hypotheses can be formed out of interaction with the world and unsupervised representation learning. Finally, active structured Bayesian inference is computationally very costly \cite{buckley2017a}, and quickly become computationally intractable for problems of moderate size. 

The challenges to address efficient and tractable curiosity-driven exploration in real world situations also underly the development of heuristic theories of curiosity-driven learning \cite{oudeyer2013a}. Within the perspective of the 'child as a sense making organism', these heuristic theories have considered  what mechanisms could enable efficient exploration and learning in high-dimensional physical bodies, and under limited resources of time, energy and cognitive capabilities. One such heuristic-oriented theoretical framework is the 'learning progress (LP) hypothesis', proposed in \cite{oudeyer2006a,kaplan2007a,oudeyer2007a,oudeyer2007b}. Here, curiosity-driven exploration in living organisms is viewed as driven by the \textbf{heuristic} estimation and search of various forms of expected learning progress\footnote{Various works in artificial intelligence, machine learning and optimal experiment design have studied, in the last 50 years, how mechanisms of active learning can push a \textit{machine} to explore parts of the state-action space that maximize forms of information gain and learning progress. However, these lines of work did not propose and study the hypothesis that related mechanisms could explain aspects of \textit{human} curiosity-driven learning, and in other animals. Coming from this modeling perspective, the LP hypothesis was developed independantly of these lines of work in machine learning and AI. The convergence of these various strands of work towards related algorithms supports the strength and scope of these mechanisms.}. More precisely, this includes a heuristic meta-cognitive mechanism that estimates expected learning progress associated to competing activities (stimuli to observe, situations or self-generated goals to engage in). This estimation of LP is then used to choose which activities to explore, selecting in priority activities that maximize expected learning progress. 

The fundamental similarity between the free-energy principle and the LP hypothesis is that both frameworks consider curiosity-driven exploration as a process that aims to collect new information to maximize the quality of an internal world model, and where this world model includes a model of self-knowledge and self-competences. Another similarity is that both frameworks consider various forms of learning progress, as the organism can learn various forms of knowledge and skills at various scales of space and time. Some forms of learning progress can result from an attentional action on a short time scale, providing information gain to better estimate the current world state. But it can be also result from the choice to focus on an activity for a longer time scale, producing various forms of improvement of a predictive world model, ranging from reduction in empirical prediction errors to reduction of uncertainty about these predictions (uncertainty could improve without improving the average prediction error, and still this would be a form of learning progress). This latter form of learning progress (longer time scale of learning, improvement of predictive world model) has been the focused of most computational experiment of the LP hypothesis so far \cite{oudeyer2016a}. 

\index{goals!autonomous goal setting} \index{goals!self-generated goals} \index{goals!curiosity-driven goal exploration}
There are also fundamental differences between the free-energy principle and the LP framework. In the free-energy framework, the mechanisms used to represent and update world models and their associated uncertainties are based on Bayesian inference. On the contrary, the LP framework has considered heuristic algorithms to learn world models from observations, and to estimate uncertainty and learning progress, using mechanisms such as memory-based and lazy learning techniques \cite{forestier2017a}, as well as neural networks \cite{kaplan2003a} and evolution strategies \cite{benureau2016behavioral}. One particular aspect of this difference is that heuristic-based approaches do not require that the learner knows all possible events, event representations, and model representations as it discovers the world (on the contrary, the normative framework requires priors about these events and their representations, which is untractable for real world situations). Unsupervised neural network representation learning techniques can for example be used to learn new spaces of representations in which to make predictions and set goals, as the world is being discovered (e.g. \cite{pere2018a}). Unsupervised learning techniques are also used in the LP framework to learn incrementally abstractions of the low-level sensorimotor flow, enabling for example to form distinct concepts of the self, of inanimate objects and of others based on their associated learnability properties \cite{oudeyer2007a}. Finally, another difference already mentionned earlier, is that the LP framework has been extended to cover mechanisms of autonomous goal setting, which is a key dimension of curiosity-driven exploration in humans. 

As the LP framework has studied what heuristics can drive efficient learning of world models in the real world, evaluation has leveraged robotics experiments under constraints of time and processing, showing how these mechanisms enable learning multiple forms of locomotion \cite{baranes2013a}, manipulation of flexible objects \cite{nguyen2014a}, and tool use discovery \cite{forestier2017a} in high-dimensional continuous action and perceptual spaces.

Beyond its theoretical origin, the learning progress hypothesis makes behavioural predictions that are compatible with a growing set of experimental evidences in psychology. For example, Gerken et al. \cite{gerken2011a} showed that 17-months old children attend more to learnable artificial linguistic patterns than to unlearnable ones. Also, Kidd et al. \cite{kidd2012a} showed that infants prefer sequences of perceptual stimuli that have an intermediate level of complexity. Similarly, Begus et al. \cite{begus2016a} showed that infants selectively ask the help of informants based on expected information they can provide. Baranes et al. \cite{baranes2014a} showed how adult subjects, who were free to explore and select tasks of various difficulties, spontaneously organize their exploration in growing order of difficulty and settle on levels of intermediate difficulty just beyond their current skill level. 

\section{Curiosity and Self-Organization in Development}
\index{self-organization} \index{development} \index{development!cognitive} \index{development!sensorimotor}

These computational studies of the learning progress hypothesis have also uncovered a crucial emergent property of such a heuristic mechanism: it spontaneously leads the learner to avoid situations which are either trivial or too complicated\footnote{Other heuristics like novelty or surprise bonuses proposed in the RL litterature do not scale as well in open real world environments as there are many tasks or situations which produce novelty or surprise but yet should be avoided as they are not learnable. An activity can be unlearnable due to unobservable causal factors, or to intrinsic unpredictability.}, and focus on situations that are just beyond the current skills in prediction or control, exploring them as long as learning progress happens in practice. 

This has enabled to generate the new hypothesis that mechanisms of curiosity drive the emergence of ordered developmental trajectories at long time scales \cite{kaplan2007a,oudeyer2016a}.
Several studies have shown that such trajectories match several fundamental properties of infant trajectories in domains such as vocal development \cite{moulin-frier2014a}, imitation \cite{kaplan2007b} and tool use discovery \cite{forestier2016b,forestier2016c}. Related models of curiosity-driven learning, with intrinsic rewards based on information gain measured through empirical prediction errors, have also shown how it could model the formation of pro-social behaviors \cite{baraglia2016a}, or model the development of aspects of binocular vision \cite{zhu2017a} or visual search in infants \cite{twomey2017a}, as well as different forms of exploratory behaviours in other animals \cite{gordon2014a}. 

\subsection{The Playground Experiment}
\index{Playground experiment} \index{curiosity!curiosity-driven learning} \index{curiosity!curiosity-driven learning robots}

An example of self-organization of structured developmental trajectories driven by curiosity-driven exploration is the Playground Experiment \footnote{The text describing the Playground Experiment in this section, and the interaction of curiosity with social guidance below, is partly adapated from \cite{oudeyer2016a}.} \cite{oudeyer2006a,kaplan2007a,oudeyer2007a}. It illustrates how mechanisms of curiosity-driven exploration, dynamically interacting with learning, physical and social constraints, can self-organize developmental trajectories. In particular, this leads a learner to successively discover two important functionalities: object affordances and vocal interaction with its peers.

\begin{figure}[!ht]
\begin{center}
\includegraphics[width=\textwidth]{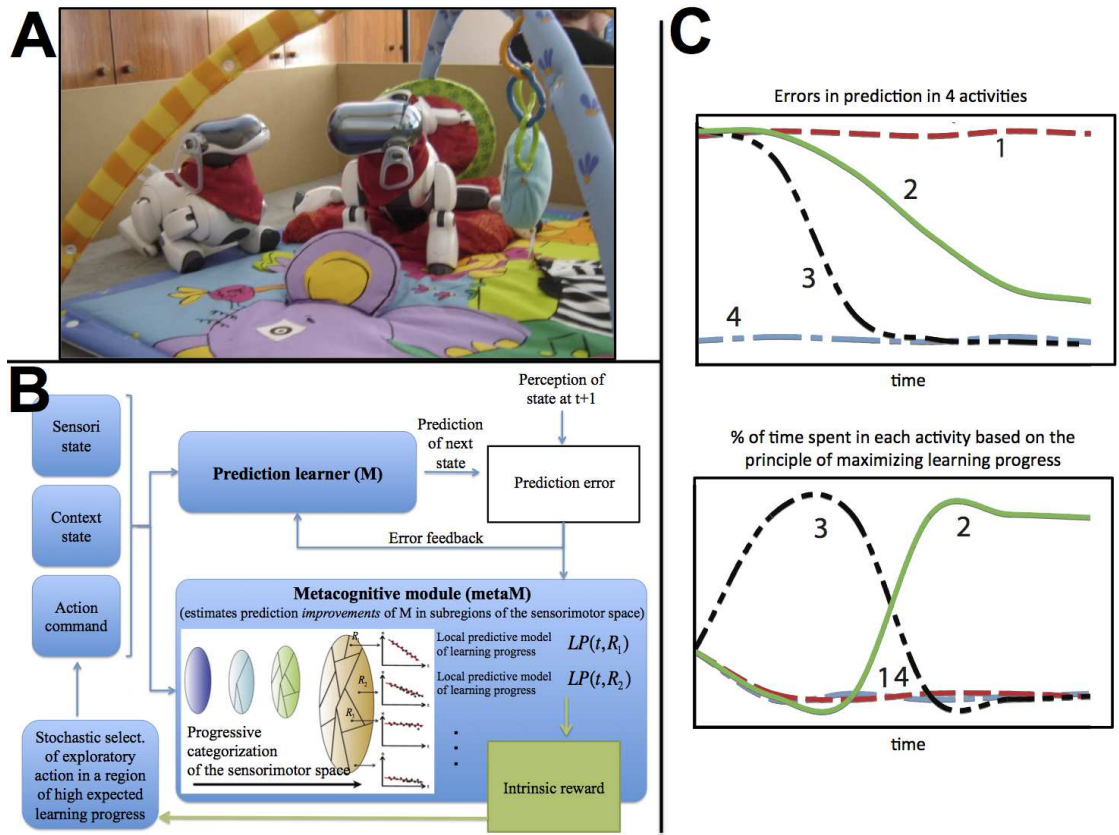}
\end{center}
\caption{The Playground Experiment \cite{oudeyer2006a,oudeyer2007a}  (A) The learning context; (B) The computational architecture for curiosity-driven exploration: 1) the robot learner probabilistically selects actions and contexts according to their potential to provide information that improves the world model (i.e. reduces prediction errors); 2) an unsupervised learning algorithms progressively differentiates actions and contexts to be selected; (C) Illustration of a self-organized developmental sequence where the robot automatically identifies, categorizes and shifts from simple to mode complex learning experiences. Figure adapted with permission from \cite{gottlieb2013a}.}
\label{fig:2}
\end{figure}

In these Playground Experiments, a quadruped “learning” robot (the learner) is placed on an infant play mat with a set of nearby objects and is joined by an “adult” robot (the teacher), see Figure~\ref{fig:2} (A) \cite{oudeyer2006a,kaplan2007a,oudeyer2007a}. On the mat and near the learner are objects for discovery: an elephant (which can be bitten or “grasped” by the mouth), a hanging toy (which can be “bashed” or pushed with the leg).  The teacher is pre-programmed to imitate (with a different pitch of voice) the sounds made by the learner when the learning robot looks to the teacher while vocalizing at the same time.

 The learner is equipped with a repertoire of motor primitives parameterized by several continuous numbers that control movements of its legs, head and a simulated vocal tract. Each motor primitive is a dynamical system controlling various forms of actions: (a) turning the head in different directions; (b) opening and closing the mouth while crouching with varying strength and timing; (c) rocking the leg with varying angle and speed; (d) vocalizing with varying pitch and length. These primitives can be combined to form a large continuous space of possible actions. Similarly, sensory primitives allow the robot to detect visual movements, salient visual properties, proprioceptive touch in the mouth, and pitch and length of perceived sounds. For the robot, these motor and sensory primitives are initially black boxes and he has no knowledge about their semantics, effects or relations. 

The learning robot learns how to use and tune these primitives to produce diverse effects on its surrounding environment, and exploration is driven by the maximization of learning progress: the robot chooses to perform physical experiences (“experiments”) that improve maximally the quality of predictions of the consequences of its actions, i.e. that improve its world model. 

Figure~\ref{fig:2} (B) outlines the computational architecture of curiosity-driven learning, called IAC, used in the playground experiment \cite{oudeyer2007a}. A prediction machine (M) learns to predict the consequences of actions taken by the robot in given sensory contexts. For example, this module might learn to predict (with a neural network) which visual movements or proprioceptive perceptions result from using a leg motor primitive with certain parameters. A meta-cognitive module estimates the evolution of errors in prediction of M in various regions of the sensorimotor space. More precisely, this module estimates the decrease in prediction errors for particular kinds of actions or particular contexts. An example of such a context could be the prediction of the consequences of a leg movement when this action is applied towards a particular area of the environment. These estimates of error reduction, measuring a form of learning progress, are used to compute an intrinsic reward. This reward is an internal quantity (a number) that is proportional to the decrease of prediction errors, and the maximization of this quantity is the objective of action selection within a computational reinforcement learning architecture \cite{kaplan2003a, oudeyer2007a,oudeyer2007b}. Importantly, the action selection system chooses most often to explore activities where the expected intrinsic reward is high, i.e. where the expected learning progress is high. However, this choice is probabilistic, which leaves the system open to learning in new areas and open to discovering other activities that may also yield progress in learning\footnote{Technically the decision on how much time to spend on a given activity/context is achieved using Multi-Armed Bandit algorithms for the so-called exploration/exploitation dilemma \cite{audibert2009a}. As the measure of learning progress in each arm is used as the reward to maximize, this is a non-stationary bandit algorithm setting. However, a specificity of learning architectures used in the robotic experiments presented here is that instead of relying on a set of pre-defined bandit arms \cite{audibert2009a}, an unsupervised learning algorithm dynamically builds new bandit arms to select from \cite{baranes2009TAMD}.}.  Since the sensorimotor flow does not come pre-segmented into activities and tasks, a system that seeks to maximize differences in learnability is also used to progressively categorize the sensorimotor space into differentiated contexts. This categorization thereby models the incremental creation and refining of cognitive categories differentiating activities/tasks. 

To illustrate how such an exploration mechanism can automatically generate ordered learning stages, let us first imagine a learner confronted with four categories of activities, as shown on figure~\ref{fig:2} (C). The practice of each of these four activities, which can be of varying difficulty, leads to different learning rates at different points in time (see the top curves, which show the evolution of prediction errors in each activity if the learner were to focus full-time and exclusively on each). If, however, the learner uses curiosity-driven exploration to decide what and when to practice by focusing on progress niches, it will avoid activities already predictable (curve 4) or too difficult to learn to predict (curve 1), in order to focus first on the activity with the highest learning rate (curve 3) and eventually, when the latter starts to reach a 'plateau', to switch to the second most promising learning situation (curve 2). Thus, embodied exploration driven by learning progress creates an organized exploratory strategy, i.e. a developmental trajecotory: the system systematically achieves these learning experiences in an order and does so because they yield (given the propensities of the learner and the physical world) different patterns of uncertainty reduction.

In the Playground experiment, multiple experimental runs lead to two general categories of results: self-organization and a mixture of regularities and diversities in the developmental patterns \cite{oudeyer2006a,oudeyer2007a}.

\subsection{Self-Organization}
In all of the runs, one observes the self-organization of structured developmental trajectories, where the robot explores objects and actions in a progressively more complex stage-like manner. During this exploration, the robot acquires autonomously diverse affordances and skills that can be reused later on and that change the learning progress in more complicated tasks, triggering a developmental cascade. The following developmental sequence was typically observed: 
\begin{enumerate}
\item In a first phase, the learner achieves unorganized body babbling;
\item In a second phase, after learning a first rough model and meta-model\footnote{The 'model' refers here to the predictive world model being learnt, which enables to predict the consequences of actions in a given context. The 'meta-model' is another model built by the meta-cognite process, and continuously estimates expected learning progress of the lower-level world model.}, the robot stops combining motor primitives, exploring them one by one, but each primitive is explored itself in a random manner;
\item In a third phase, the learner begins to experiment with actions towards zones of its environment where the external observer knows there are objects (the robot is not initially provided with a representation of the concept of “object”), but in a non-affordant manner (e.g. it vocalizes at the non-responding elephant or tries to bash the teacher robot which is too far to be touched);
\item In a fourth phase, the learner now explores the affordances of different objects in the environment: typically focussing first on grasping movements with the elephant, then shifting to bashing movements with the hanging toy, and finally shifting to explorations of vocalizing towards the imitating teacher.
\item In the end, the learner has learnt sensorimotor affordances with several objects, as well as social affordances, and has mastered multiple skills. None of these specific objectives were pre-programmed. Instead, they self-organized through the dynamic interaction between curiosity-driven exploration, statistical inference, the properties of the body, and the properties of the environment.
\end{enumerate}

These playground experiments do not simply simulate particular skills (such as batting at toys to make them swing or vocalizations) but simulate an ordered and systematic developmental trajectory, with a universality and stage-like structure that may be mistakenly taken to indicate an internally-driven process of maturation.  However, the trajectory is created through activity and through the general principle that sensorimotor experiences that reduce uncertainty in prediction are rewarding.  In this way, developmental achievements can build on themselves without specific pre-programmed dependencies but nonetheless – like evolution itself –create structure  (see \cite{smith2007a,smith2013a}, for related findings and arguments). 

\subsection{Regularities and Diversity}
Because these are self-organizing developmental processes, they generate not only strong regularities but also diversity across individual developmental trajectories. For example, in most runs one observes successively unorganized body babbling, then focused exploration of head movements, then exploration of touching an object, then grasping an object, and finally vocalizing towards a peer robot (pre-programmed to imitate). This can be explained as a gradual exploration of new progress niches, and those stages and their ordering can be viewed as a form of attractor in the space of developmental trajectories. Yet, with the same mechanism and same initial parameters, individual trajectories may invert stages, or even generate qualitatively different behaviours. This is due to stochasticity (the same motor commands do not produce always the same results), to small variability in the physical realities and to the fact that this developmental dynamical system has several attractors with more or less extended and strong domains of attraction (an attractor is a part of the state-space in which the dynamical system converges, depending on what was his initial state).  We see this diversity as a positive outcome since individual development is not identical across different individuals but is always, for each individual, unique in its own ways. This kind of approach, then, offers a way to understand individual differences as emergent in developmental processes itself and makes clear how developmental processes might vary across contexts, even with an identical learning mechanism.

A further result to be highlighted is the early development of vocal interaction.  With a single generic mechanism, the robot both explores and learns how to manipulate objects and how to vocalize to trigger specific responses from a more mature partner \cite{oudeyer2006a,kaplan2007a}. Vocal babbling and language games have been shown to be key in infant language development; however, the motivation to engage in vocal play has often been associated with hypothesized language specific motivation. The Playground Experiment makes it possible to see how the exploration and learning of communicative behaviour might be at least partially explained by general curiosity-driven exploration of the body affordances, as also suggested by Oller \cite{oller2000a}. Exploring this idea further, Forestier and Oudeyer \cite{forestier2017b} studied simulation showing how these mechanisms can drive the joint development of speech and tool use, where speech is discovered as a particular tool enabling to get social peers achieve targeted actions.

\subsection{Interaction with Social Guidance}
\index{curiosity!and social guidance}
Other robotic models have explored how social guidance can be leveraged by an intrinsically motivated active learner and dynamically interact with curiosity to structure developmental trajectories \cite{thomaz2008a,nguyen2014a}. Focusing on vocal development, Moulin-Frier et al. conducted experiments where a robot explored the control of a realistic model of the vocal tract in interaction with vocal peers through a drive to maximize learning progress \cite{moulin-frier2014a}. This model relied on a physical model of the vocal tract, its motor control and the auditory system. The experiments showed self-organization of vocal development trajectories that share structural similarities with infants \cite{oller2000a}. They showed how these mechanism generate an adaptive transition from vocal self-exploration with little influence from the speech environment, to a later stage where vocal exploration becomes influenced by vocalizations of peers. Within the initial self-exploration phase, a sequence of vocal production stages self-organizes, and shares properties with infant data: the vocal learner first discovers how to control phonation, then vocal variations of unarticulated sounds, and finally articulated proto-syllables. As the vocal learner becomes more proficient at producing complex sounds, the vocalizations of the teacher become vocal goals to imitate that provide high learning progress, resulting in a shift from self-exploration to vocal imitation.

\section{Challenges and Perspectives}

Computational theories have enabled to better understand the potential structures and functions of curiosity-driven learning in the last decade. These theories have identified a wide diversity of algorithmic mechanisms that could produce the kind of spontaneous exploration displayed by humans and other animals. This diversity concerns both the measures of interests (e.g. novelty, surprise, learning progress, knowledge gap, intermediate complexity, ...) and the entities to which the brain may apply them (e.g. actions, states, goals, objects, tools, places, games, activities, learning strategies, social informants, ...), with time scales ranging from the moment-to-moment to days and months. Furthermore, theoretical models of curiosity-driven learning, and their application in artificial intelligence and machine learning, have shown the key role of these mechanisms for making discoveries and solving real-world problems with rare or deceptive rewards, in large and changing environments. In brief, computational theories:

\begin{enumerate}
	\item have shown that the term 'curiosity' covers a wide diversity of complex mechanisms, generating different forms of exploration; 
	\item have proposed ways to model and study these mechanisms formally, contributing to the naturalization of the concept of 'curiosity'; 
	\item have shown that curiosity mechanism are essential to learning and development, and thus should become a central topic in cognitive sciences and neurosciences.
\end{enumerate}

Related work in psychology and neuroscience are beginning to converge with computational theories towards conceptualizing how mechanisms of curiosity can play a fundamental role in many aspects of development, ranging from sensorimotor, cognitive, emotional, to social development. However, for multiple reasons, experimental work studying the underlying mechanisms of curiosity have been very limited so far in psychology and neuroscience. The empirical testing of computational theories have been for a large part beyond the reach of existing experimental paradigms in psychology and neuroscience. Several challenges need to be addressed to leverage further the interaction between theory and experimentation.

\index{experimental paradigms}
\paragraph{The need for novel experimental paradigms in psychology and cognitive neuroscience.} 
A first general challenge is that curiosity denotes a set of mechanisms that push individuals to explore what is interesting for themselves, out of the consideration of external tasks or expectations of social peers. Yet, the very act of participating to an experiment in a lab brings up expectations in the subject's mind about what the experimenter wants to observe or analyze, or will think about what they do. In the lab, curiosity can disappear quickly as soon as one begins to observe it. This is probably less the case with very young infants, but in their case the presence of social peers is also bound to influence what they do, and their limited capabilities for physical exploration and verbal reporting makes it difficult to study advanced forms of curiosity. So, how to study curiosity when setting up a controlled experiment introduces complex interaction with other motivational forces that are hard to control and evaluate? It is interesting to note that the most clear observations of curiosity in the lab do not come from studies targeting curiosity and information-seeking, but are rather observations of child behaviour spontaneously doing things that are wildly different from the task the experimenter designed for them. For example, in recent experiments of Lauriane Rat-Fisher and colleagues\footnote{personal communication} about tool use development, children are asked to retrieve a salient toy stuck in a tube (the toy was expected to be very attractive to the child). Yet, several children showed spontaneous strong intrinsic interest in exploring how to push sticks and objects in the tube, continuing to do it with a lot of fun after getting the toy out of the tube, and completely ignoring the toy. Unfortunately, these “making off” observations are typically removed from the lens of analysis of traditional experimental studies, while they may display some of the most fundamental and mysterious mechanisms of learning and cognition. 

Another experimental challenge is how to disentangle the potentially different mechanisms identified by theory, which may be simultaneously at play in individuals, and potentially on different time scales. For example, it could be possible that curiosity-driven attention on very short time scales may be driven by intrinsic rewards measuring different forms of novelty, surprise or prediction error. However, on longer time-scales, the curious brain may value the intrinsic interest of activities, games or goals with other measures of interest like learning progress. The study of curiosity over long time scales, focusing on how it may contribute to sculpt sensorimotor, cognitive and social development, on how it develops itself with time, and on how it interacts with other developmental forces such as social learning, is maybe the most important and most difficult challenge in this scientific area.  

\section{Acnknowledgements}
This manuscript has benefited from very useful feeback and discussions with members of the Flowers team at Inria,
as well as with Jacqueline Gottlieb, Linda Smith and Olivier Sigaud. 

\bibliographystyle{apalike}
\bibliography{oudeyer}

\bigskip

\label{lastpage-01}

\end{document}